\documentclass[lettersize,journal]{IEEEtran}
\usepackage{multido}
\usepackage{pgffor}
\usepackage{graphicx} 
\usepackage{caption}
\usepackage{algorithm}
\usepackage{subcaption}
\usepackage{float}
\usepackage{mathtools}
\usepackage{amsmath}
\usepackage{algorithm}
\usepackage{algpseudocode}
\usepackage{tikz}
\usepackage{amssymb}

\newcommand{\vy}{\vec{\itbf{y}}}

\newcommand{\vx}{\vec{\itbf{x}}}

\DeclareMathAlphabet{\itbf}{OML}{cmm}{b}{it}

\usepackage{mathrsfs}
\usepackage{algpseudocode}
\usepackage{bbm}
\usepackage{amsthm}
\theoremstyle{remark}

\usepackage{amsmath,amssymb,amsfonts}

\usepackage{tikz}
\usepackage{pgfplots}
\usepgfplotslibrary{groupplots}
\usepackage{tikz}
\usetikzlibrary{positioning}
\usetikzlibrary{arrows.meta, bending, chains, decorations.markings}

\newcommand{\C}{\mathbb{C}}

\algrenewcommand\algorithmicrequire{\textbf{Input:}}
\algrenewcommand\algorithmicensure{\textbf{Output:}}

\usepackage{siunitx}
\usepackage{hyperref}
\hypersetup{
    colorlinks=true,
    linkcolor=blue,
    filecolor=magenta,      
    urlcolor=cyan,
    }
\usepackage{xcolor}
\usepackage{comment}

\usepackage{dcolumn}
\usepackage{bm}
\usepackage{amssymb,amsfonts,amsmath}
\usepackage[normalem]{ulem}

\def\bfX{{\bf X}}
\def\bfY{{\bf Y}}
\def\bfD{{\bf A}} 
\newcommand{\scatt}{\boldsymbol{\xi}}
\newcommand{\psource}{\mathbf{y}}
\newcommand{\pos}{\mathbf{x}}
\newcommand{\winc}{\hat{\psi}^{i}}
\newcommand{\wscatt}{\hat{\psi}^{s}}
\newcommand{\wexc}{\hat{\psi}^{e}}

\def\bfv{{\bf v}}

\def\Gc{{\cal G}}

\def\vect#1{\mbox{\boldmath{$#1$}}}

\def\vx{{\vec z}}
\def\psource{{\vec z}}

\def\vy{{\vec r}}
\def\pos{{\vec r}}
\newcommand{\bxx}{\vect x}
\newcommand{\byy}{\vect y}

\def\mC{\mathbb{C}}

\def\wg{{g}}

\newcommand\ones{1\kern-0.25em\text{l}}

\graphicspath{{../FoldyPRL/submit/}}

\begin{document}



\title{Super-resolution in disordered media using neural networks}
\author{Alexander Christie\IEEEauthorrefmark{1}, Matan Leibovich\IEEEauthorrefmark{2}, Miguel Moscoso\IEEEauthorrefmark{3}\\ Alexei Novikov\IEEEauthorrefmark{4}, George Papanicolaou\IEEEauthorrefmark{1}, Chrysoula Tsogka\IEEEauthorrefmark{5}

\thanks{\IEEEauthorrefmark{1}{Stanford University, Stanford, CA 94305, USA}}
\thanks{\IEEEauthorrefmark{2}{Courant Institute of Mathematical Sciences, New York University, New York, NY 10012, USA}}
\thanks{\IEEEauthorrefmark{3}{Universidad Carlos III de Madrid, Leganes, Madrid 28911, Spain}}
\thanks{\IEEEauthorrefmark{4}{The Pennsylvania State University, University Park, PA 16802, USA}}
\thanks{\IEEEauthorrefmark{5}{University of California, Merced, CA 95343, USA}}}

\markboth{}%
{\MakeLowercase{\textit{et al.}}: Super-resolution in disordered media using NN}

\maketitle

\begin{abstract}
We propose a methodology that exploits large and diverse data sets to accurately estimate the ambient medium's Green's functions in strongly scattering media.  Given these estimates, obtained with and without the use of neural networks, excellent imaging results are achieved, with a resolution that is better than that of a homogeneous medium. This phenomenon, also known as super-resolution, occurs because the ambient scattering medium effectively enhances the physical imaging aperture.
\end{abstract}


{\em This work has been submitted to the IEEE for possible
publication. Copyright may be transferred without notice, after
which this version may no longer be accessible} 
\maketitle


\section{Introduction}
\IEEEPARstart{I}{maging} in strongly scattering media is challenging because the multiple scattering leads to loss of directional information and coherence. 
Unless the propagation medium is known, or can be accurately estimated, coherent imaging typically fails in such media as it relies on using some approximation of the unknown background's Green's function. In scattering media, coherent interferometry (CINT) may be used successfully \cite{BPT06, BGPT11}, providing statistically stable results by back propagating cross-correlations of the data over adaptively chosen space and frequency windows, their size being determined by the coherence of the data. In strongly scattering media, the coherence of the data is quite low, which severely limits the imaging resolution of CINT. 

The methodology we propose in this paper allows for imaging in strongly scattering media with super-resolution. The enhanced resolution 
arises from an effective array aperture, larger than the physical array, as a result of scattering \cite{Fink2000,BPTB02,let17}. The improvement in imaging resolution is achievable because the Green's functions for propagation between the array and the imaging region are accurately estimated by the proposed methodology, using both  conventional optimization algorithms and neural networks. The neural network approach is more flexible, since it does not require any initialization, while an appropriate one is necessary for using the other algorithms. 

Motivated by \cite{MNPT_PNAS}, which deals with weakly scattering media, our approach for estimating the array Green's function vectors in a strongly scattering medium without the use of neural networks involves three steps. First, using correlation-based methods we determine a set of representative Green's function vectors, 
or a dictionary, such that every array data vector is a linear combination of a small number of them. This algorithm, developed in \cite{NW}, 
relies on the assumed sparsity of sources in the data vectors and the availability of a diverse 
and abundant array data set, but does not assume any prior knowledge of source locations or amplitudes.
The representative vectors obtained from this algorithm 
are a good estimate of the columns of the sensing matrix. 
Each of these columns is a Green's function vector between a focal or source point in the image window and the receiver array.
The output of this first step is limited since
the estimated columns are (a) not accurate enough, and (b) unordered.
Ordering the columns of the sensing matrix is crucial because in order to generate an image we need to know which source point in the image window corresponds to which column of the sensing matrix.

In the second step, we improve the accuracy of the estimated columns by the Method of Optimal Directions, an optimization-based algorithm \cite{Engan00,agarwal}. Since the algorithm is non-convex a good enough initialization is required, which is provided by step one.

The algorithm outputs a much more accurate estimate of the sensing matrix. However, it is still unordered.
These two steps are an accurate dictionary learning approach for the unordered columns of the sensing matrix. 

As an alternative to the sparse dictionary learning/ non-convex optimization, we also propose a two-step neural network approach. The two steps in the neural network approach are (a) training an encoder-decoder neural network with unlabeled array data to obtain several versions of accurate, unordered columns of the sensing matrix, plus a few few inaccurate estimates, to be discarded, and (b) using a clustering algorithm on the combined set of all versions of the columns of the sensing matrix from step (a) to obtain an accurate estimate of each column of the sensing matrix. We provide a comparison of the two approaches along with an assessment of their relative advantages and disadvantages.

Following the stimation of the unordered sensing matrix, we order its columns using a connectivity-based multidimensional scaling algorithm \cite{Shang03,Oh10}, relying on cross correlations to associate each column of this matrix with its corresponding focal point in the image window, in a similar manner to \cite{MNPT_PNAS}. 
 This determines the correct order of columns of the estimated matrix,  which is essential for generating images.

Once the ordered sensing matrix has been estimated, very accurate images can be obtained with commonly used $L^2$ methods like migration, or sparsity promoting  $L^1$ imaging methods. Here we obtain an image by migration, that is,
by applying the conjugate adjoint of the estimated sensing matrix to any array data vector to obtain the corresponding image. The resulting resolution is much better than that of a homogeneous medium with the same imaging setup. Thus, we achieve super-resolution, similar to resolutions achieved by physical time reversal \cite{Fink2000,BPTB02,let17}. However, while physical time reversal is not imaging because we have no way of knowing where the sharply focused points, the sources, are located, the proposed methodology allows us to locate them very accurately, and so obtain super-resolution images.

The proposed approach generalizes directly to the scattering case, that is, when small scattering objects are present in the imaging domain rather than sources. This is done by using the effective source formalism introduced in \cite{Moscoso14}.

The main result of this paper is a demonstration of the feasibility of the above three-step estimation algorithm, with and without the use of neural networks, for achieving super-resolution imaging. We support this with numerical simulations in the microwave C-band radar regime. 

We formulate the imaging problem next, including the Foldy-Lax model for simulating a strongly scattering medium, as shown in Figure \ref{fig:schema}. In Section \ref{sec:3step} we present the three step optimization based method for imaging, without using neural networks.
In Section \ref{s:numersim} we present the numerical simulations in the C-band radar regime with Figure \ref{fig:image} showing super-resolution, which is the main result
of this paper. In Section \ref{sec:NN} we introduce the neural network approach to imaging with super-resolution in strongly scattering media. Finally 
Section \ref{sec:sumc} is a summary with some conclusions and a brief comparison of the conventional optimization approach to that of using neural networks.

\section{The Imaging problem} 
An array of $N$ receivers records waves generated by sources located in a region of interest, the image window. 
The direction parallel to the array orientation is the cross-range and the one orthogonal to it is the range. 
The Green's function characterizes the propagation of a signal of frequency $\omega$ from a point $\vx$ 
to a point $\vy$ in the scattering medium, and satisfies the wave equation of the medium. We model the scattering medium, using the Foldy-Lax equations~\cite{foldy,lax,let17,martin}.  This model represents a discrete, multiple scattering medium, that is, a collection of small scatterers distributed randomly in an otherwise uniform background.

\subsection{The Foldy-Lax model}\label{ss:FL}
Multiple scattering using the Foldy-Lax equations in the frequency domain is as follows. Let $\winc(\pos, \psource) = G_{0}(\pos - \psource)$ be the incident wave at position $\pos$ due to a source at position $\psource$,
where
\begin{equation}
  G_{0}(\pos - \psource) : =  \frac{e^{i k |\pos - \psource|}}{4 \pi
    |\pos - \psource|},
\label{eq:incwave}
\end{equation}
and where $k=\omega/c_0$ is the wavenumber with $c_0$ the constant, background wave speed. 
The total wave $\hat{\psi}(\pos, \psource)$ at $\pos$ due to a source at $\psource$ can be written as 
the sum of the incident and scattered waves:
\begin{equation}
  \hat{\psi}(\pos, \psource)= 
  \winc(\pos, \psource)+ \sum_{j=1}^{J}
  \wscatt_j(\pos, \psource).
\label{eq:totalwave}
\end{equation}
Here, $\wscatt_{j}(\pos,\psource)$ is the wave arising from a scatterer at position
$\scatt_j$, evaluated at position $\pos$. 
It is given by
\begin{equation}
  \wscatt_j(\pos,\psource) =
  G_{0}(\pos - \scatt_{j}) \tau_j \wexc_j(\psource),
  \label{eq:asymp}
\end{equation}
 with $\tau_{j}$ denoting the scattering
amplitude, and $\hat{\psi}_{j}^{e}$ denoting the exciting wave at the
scatterer located at $\scatt_{j}$.

We ignore self-interacting waves so that the exciting wave
$\hat{\psi}^{e}_{j}$ is equal to the sum of the incident wave
$\hat{\psi}^{i}$ at $\scatt_j$ and the scattered waves at $\scatt_j$
due to all scatterers except for the one at $\scatt_{j}$. 
Assuming that the scatterers are sufficiently far apart from
each other the wave $\hat{\psi}^{e}_{j}$ is given by
\begin{equation}
  \wexc_j ( \psource)=
  \winc(\scatt_j, \psource) +
  \sum_{m = 1,\ m \neq j}^{J} G_{0}(\scatt_{j} - \scatt_{m}) \tau_{m} 
  \wexc_{m}(\psource).
\label{eq:scattwave}
\end{equation}
Eq.~\eqref{eq:scattwave} is a self-consistent system of $J$ equations
for the $J$ unknown  exciting waves $\wexc_{1}, \dots, \wexc_{J}$.
This system can be written in matrix form and then solved numerically by
standard methods. Once the exciting waves have been computed, we find 
the total wave  $\hat{\psi}(\pos, \psource)$ at $\pos$ due to a source at $\psource$ using Eqs.~\eqref{eq:totalwave} and \eqref{eq:asymp}.

%

\begin{figure}[t!]
\begin{center}
\begin{minipage}{0.9\linewidth}
 \includegraphics[width=0.9\linewidth]{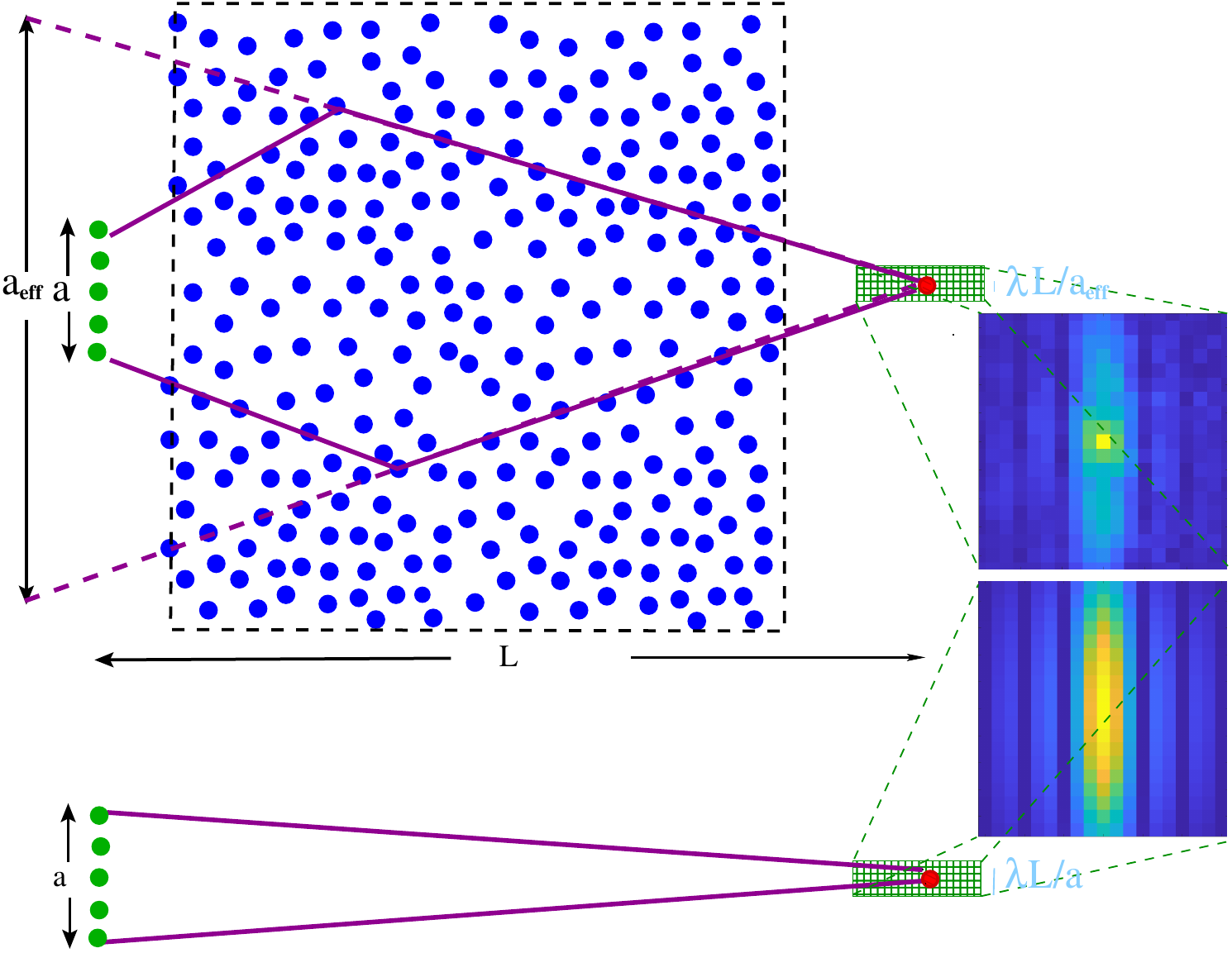}
\end{minipage}
\end{center}
\caption{\label{fig:schema} \textbf{Imaging schematic and illustration of the effective aperture}. 
In the simulations, the cross-range resolution in homogeneous media is $\lambda L/a=8\lambda$, where $\lambda$ is the central wavelength, $a$ is the physical array aperture and $L$ is the range. The
range resolution is $c_0/B$ where $c_0$ is the background propagation speed and $B$ is the bandwidth.
In the scattering medium the resolution is about $2\lambda$ which translates to an effective aperture 
that is four times the physical aperture of the array, {\em i.e.,} $a_{eff} \approx 4 a = 120 \lambda$. The range resolution in the scattering medium does not change.}
\end{figure}

The simulation setup is illustrated in Figure~\ref{fig:schema}. Multiple scattering in disordered media leads to a larger effective aperture as illustrated on the top panel of Figure~\ref{fig:schema}, which, in turn, leads to  
 super-resolution, discussed below in detail.

We note that the increase in the effective aperture size is the result of increased cross range diversity. The weakly scattering wave propagation model used in~\cite{MNPT_PNAS} does not provide cross-range diversity, so that the scattered waves are not distributed over a larger part of the domain than they would be in a homogeneous medium. In other words, weakly scattering media
do not give rise to an effective array aperture larger than the physical one due to multipathing. On the other hand, the Foldy-Lax model does generate cross-range diversity and, therefore, a larger effective aperture that captures more information about the location of the sources, which is the direct cause of super-resolution.

We note that super-resolution is observed numerically in more complex multiple scattering models, such as in \cite{let17},  where finite size particles are considered rather than point scatterers, using a spherical harmonic expansion and a fast multipole algorithm for solving the resulting system of equations.

\subsection{Imaging setup}
We denote the Green's function vector 
\begin{equation}\label{eq:GreenFuncVec}
\vect \wg(\vx)=[ \hat{\psi}(\vy_{1},\vx),  \hat{\psi}(\vy_{2},\vx),\ldots,
 \hat{\psi}(\vy_{N},\vx)]^{T}\,,
\end{equation}
which is the data recorded on an array of $N$ receivers, located at points $\vy_{1},\dots, \vy_{N}$, arising from a single source , located at $\vx$. Here $\hat{\psi}(\pos_i, \psource)$ is the total field at $\pos_i$, $i=1,\ldots,N$, due to a source at $\psource$ obtained with the Foldy-Lax equations.

This vector is a column of the $N \times K$ sensing matrix
\begin{equation}\label{eq:sensingmatrix}
\vect{\Gc} =[\vect\wg(\vx_1)\,\cdots\,\vect\wg(\vx_K)] \ ,
\end{equation}
defined on a grid $\{\vx_k\}_{k=1,\dots,K}$  spanning the image window, with $K>N$, typically.
This grid is chosen so that it is compatible with an expected image resolution and can be adjusted empirically to the data and to the imaging system.
In effect, the sensing matrix $\vect{\Gc}$
maps a distribution of sources in the image window to 
data received on the array. That is, for a given configuration of sources on the grid, represented by the vector $\bxx\in \mC^{K}$, 
the data recorded on the array is given by
\begin{equation}
\label{eq:model}
\byy = \vect{\Gc}\, \bxx \, ,
\end{equation}
where $\byy \in \mC^{N}$.
We consider $\bxx$ as the image, that is, a vector whose $k$-th component represents the complex amplitude of the source at location $\vx_k$ in the image window, ${k=1,\dots,K}$. 

When several frequencies are involved, the array data for each
frequency are stacked into a single data vector. For simplicity, we describe the
single frequency case, but use multi-frequency data in the numerical simulations.

Because the locations of the scatterers in the medium 
are random and not known, the sensing matrix $\vect{\Gc}$ in \eqref{eq:model}  is also random and not known. In order to exploit super-resolution for imaging we need a good estimate of this matrix. To this end, we assume that we obtain $M$ diverse samples or observations $\{\byy_i\}_{i=1,\dots,M}$, with $\byy_i = \vect{\Gc}\, \bxx_i$. 
The number of observations is large with respect to the dimension of the vectors $\bxx_i$, i.e., $M \gg K$. While the $\bxx_i$'s are unknown, we assume that they are sparse, 
implying that the samples  $\byy_i$ can be represented as a linear combination of a small number of columns of the unknown sensing matrix $\vect{\Gc}$.  In the next section, we describe next the three-step estimation algorithm for imaging with super-resolution.

\section{The three-step imaging method}
\label{sec:3step}
We briefly present the three steps of our method: The first two steps of our imaging method use dictionary learning algorithms and allow us to estimate the columns of the sensing matrix $ \hat{\vect{\Gc}}$ 
up to a random permutation. Although in most applications of dictionary learning the order of the columns is not an issue, it is essential in imaging since it is not enough to accurately estimate the columns of $ \hat{\vect{\Gc}}$. Rather we need to find a consistent mapping between grid points in the image window $\{\vx_k\}_{k=1,\dots,K}$ and the $K$ different columns of $\hat{\vect{\Gc}}$. This is challenging since the propagation medium is unknown. However, we know that the estimated column vectors arise from a single focal or source point in the image window. We explain next how to identify all these focal points simultaneously in a self-consistent way.

The key idea of the third step is to use the cross-correlations between the columns of the estimated matrix $\hat{\vect{\Gc}}$ to identify nearest neighbors of their associated focal or source points in the image window, and use that to infer their associated focal points on the grid.
The grid reconstruction problem is solved using the multidimensional scaling (MDS) algorithm with a non-metric, connectivity-based distance, as in sensor network localization problems \cite{Oh10}. The proxy distance can be obtained using the normalized inner products of the estimated columns of $\hat{\vect{\Gc}}$, which are cross-correlations of Green's functions.

The cross-correlations formed can be interpreted as time reversal experiments \cite{Fink2000}. That is, the signals recorded on the array are re-emitted in the same medium and, given the time-reversibility of the wave equation, those signals will focus at the location from which the original signal was emitted. In that case, computing the cross correlation with a nearest neighbor is equivalent to using it for time-reversal, which would yield partial focusing.  Our connectivity reconstruction relies on this fundamental property of the wave equation and, therefore, is robust to the complexity of the medium. Our numerical simulations confirm that this is the case.

\subsection{Step 1: Initial estimate of the sensing matrix by dictionary learning} 
The initial estimate of the {\it unordered} columns of the sensing matrix is obtained using the correlation-based sparse dictionary learning of \cite{NW}. Sparse dictionary learning is the problem of simultaneously recovering a matrix $\bfD \in \mC^{N\times K}$ 
 and  s-sparse vectors $\bxx_i \in \mC^{K}$ from $N$ dimensional data vectors of the form 
 $\byy_i = \bfD \, \bxx_i$,   $i=1,\dots,M$. The dictionary $\bfD \in \mC^{N\times K}$ represents an estimate of the normalized sensing matrix $\vect{\Gc}$ from (\ref{eq:sensingmatrix}), up to permutations of its columns. 
 Normalized means having columns of unit length. We will use $\hat{\vect{\Gc}}$ for the ordered estimate of ${\vect{\Gc}}$. In order to find a good estimate of the columns of $\bfD$ we use the correlation-based sparse dictionary learning 
 algorithm, developed in~\cite{NW}.   Instead of seeking to recover individual columns of $\bfD$, 
 as is done in other algorithms, the method~\cite{NW} employs a spectral method to recover the subspaces spanned 
 by the columns in the support of each imaging measurement. 
  The set of subspaces is then clustered into collections of 
 subspaces that share a common column of $\bfD$.  In other words, this approach shifts the focus from identifying 
 individual columns to detecting clustering of subspaces that arise as linear combinations of those columns.  
 The individual columns are then found as intersections of these subspaces.
 This clustering part here is different from  the DBSCAN~\cite{Ester1996} method, applied later in our Neural Network approach in Section~\ref{sec:NN}. 
 Under a suitable random model of the imaging measurements, the algorithm of \cite{NW} recovers $\bfD_0$, a good estimate of columns of $\bfD$, 
 in polynomial time in $K$ and for sparsity $s$, the size of the support of the vectors $\bxx_i$, 
 growing at most linearly in $N$, up to log factors. 
 
 The necessity of this first step is the main difference between the method presented here and the one in~\cite{MNPT_PNAS}, where we investigated imaging in {\it weakly} scattering media. The first step was not needed there because it turns out 
 that  the sensing matrix of the homogeneous medium already gives a sufficiently 
 good estimate of the columns of $\bfD$ to be used in step two.

\subsection{Step 2: Improved estimate of the sensing matrix by optimization} 
\label{sec:nonc}
Define the sparse source matrix $\bfX=[\bxx_1,\dots,\bxx_M] \in \mC^{K\times M}$, and the data matrix  
$\bfY=[\byy_1,\dots,\byy_M] \in \mC^{N\times M}$.
To improve the accuracy of the dictionary $\bfD$, we 
 solve the optimization problem 
\begin{equation}
\label{dl_l1}
\begin{array}{rrclcl}
\displaystyle \min_{\bfD, \bfX} & \multicolumn{3}{l}{\| \bfD \bfX - \bfY\|_F^2 }\\
\textrm{s.t.} & \|\bxx_i\|_{0}  \leq s, \,i=1,\dots,M, \\
\end{array}
\end{equation}
where $ \|\cdot\|_0$ is the number of non-zero elements, $s$ is the expected sparsity level, and $F$ denotes the Frobenius norm.

In order to solve~\eqref{dl_l1}, we alternate between  an $\ell_1$-norm minimization problem   and a least-squares problem  to update $\bfX$ and $\bfD$ iteratively. 
Specifically, if $\bfD$ is assumed to be known, then $\bfX$ 
can be obtained from the  $\ell_1$-norm, sparsity promoting minimization problem
\begin{equation}
\label{l1}
\displaystyle \min \|{\bfX}\|_1 \,\, \textrm{ subject to } \bfD \bfX = \bfY\,.
\end{equation}
We solve \eqref{l1} using a Generalized Lagrangian Multiplier Algorithm (GeLMA), \cite{Moscoso12}.
If $\bfX$ is known, the minimization problem for $\bfD$
\begin{equation}
\label{dict}
\begin{array}{rrclcl}
\displaystyle \min_{\bfD} & \multicolumn{3}{l}{\| \bfD \bfX - \bfY\|_F^2 }\, ,\\
\end{array}
\end{equation}
can be easily solved. The exact solution for $K> N$ is $\bfD=\bfY\bfX^T(\bfX\bfX^T)^{-1}$,
provided $\bfX\bfX^T$ is invertible.
Once $\bfD$ has been computed we normalize its columns to one. The iteration~\eqref{l1}-\eqref{dict}  starts with an initial guess $\bfD_0$, for which we use the outcome of the first step - $\bfD_0$ is found by the correlation-based sparse Dictionary Learning algorithm~\cite{NW}.

At the end of the second step we obtain a dictionary matrix $\bfD$ whose columns are accurate approximations of the sensing matrix  $\vect{\Gc}$, but are unordered. The next step addresses their ordering. 

\subsection{Step 3: Grid reconstruction with multidimensional scaling} 
\label{sec:MDS}
We now describe an algorithm for finding the focal spots in the image window from the {\em estimated} Green's function vectors $\{\hat{\vect g}_i\}_{i=1}^{K}$. It is the range-free or connectivity based sensor localization algorithm \cite{Shang03}, 
analyzed in \cite{Oh10}. It is also used in \cite{MNPT_PNAS} for a weakly inhomogeneous random medium. In a strongly scattering medium, getting the connectivity information needed is challenging because the cross-correlations of the estimated, normalized, Green's function vectors $\{\hat{\vect g}_i\}_{i=1}^{K}$ must be sufficiently stable so as to allow it. In the numerical experiments we show that to attain this stability we require a large enough bandwidth and a large enough physical array in the imaging setup, even though resolution is determined by the effective aperture and not the physical one.
For each column vector $\{\hat{\vect g}_i\}$ we form its inner products, or its cross-correlations, with all other $\{\hat{\vect g}_j, j\neq i\}$,
\begin{equation}
c_{ij}=\langle \hat{\vect g}_i,\hat{\vect g}_j\rangle,
\end{equation}
and select the four whose correlations are closest to one, which are its neighbors in a two-dimensional image window that we use. We then form a graph whose nodes are identified with the $\{\hat{\vect g}_i\}$ and its edges connect nearest neighbors.
The proxy distance between two Green's function vectors is now the geodesic graph distance between their corresponding vertices as in \cite{MNPT_PNAS}. That is, the proxy distance between $ \hat{\vect g}_i$ and $\hat{\vect g}_j$, 
denoted by $\hat{d}_{ij}$, is the number of edges in the shortest path connecting $i$ and $j$. 
We use this proxy distance as a replacement of the Euclidean distance between pairs of focal points in the image window associated with Green's function vectors in the MDS algorithm  \cite{Shang03} .
The resulting configuration of focal points  $\hat{Z}=[\hat{\vx}_1, \hat{\vx}_2,\dots, \hat{\vx}_K]^T$ in the image window provides an estimate for the true 
configuration of focal points $Z=[\vx_1, \vx_2,\dots, \vx_K]^T$, up to rotation, translation and overall scaling, and hence provides the same estimate for the sensing matrix columns.

\section{Numerical simulations} \label{s:numersim}
Numerical experiments are conducted in the C-band radar regime using the Foldy-Lax model \cite{foldy,lax,let17,martin}.  
We use an imaging system operating at a central frequency of $f_0=5$GHz, wavelength $\lambda=6cm$, bandwidths $B=0.75, 1, 1.25$GHz,
 and a frequency resolution of $df=20, 20, 25$MHz, respectively. 
The imaging window dimensions are defined with a cross-range spacing of $dl_x = 0.5 \lambda$ and a range spacing of $dl_z=1.66 \lambda$, resulting in a total of $K=361$ pixels ($nx=nz=19$). 
Within a scattering area of 10$\times$10 meters, 
$M=400$ point scatterers are distributed randomly. 
An array consisting of transducers with distance $\lambda$ between them, with array size of $20,30,40 \lambda$,
is positioned at a distance of 14 meters from the center of the imaging window. Resolution in the homogeneous medium case is estimated at $\lambda L/a \approx 8 \lambda$ in cross-range 
and $5\lambda$ in range ($c_0/B$). 
For the dictionary learning step, we randomly position three point sources (sparsity $s=3$) in the image window.  We repeat the process 5000 times to generate source configurations with different random positions and strengths.
\begin{figure}[htbp]
\begin{center}
\begin{minipage}{\linewidth}
  \includegraphics[width=0.31\linewidth]{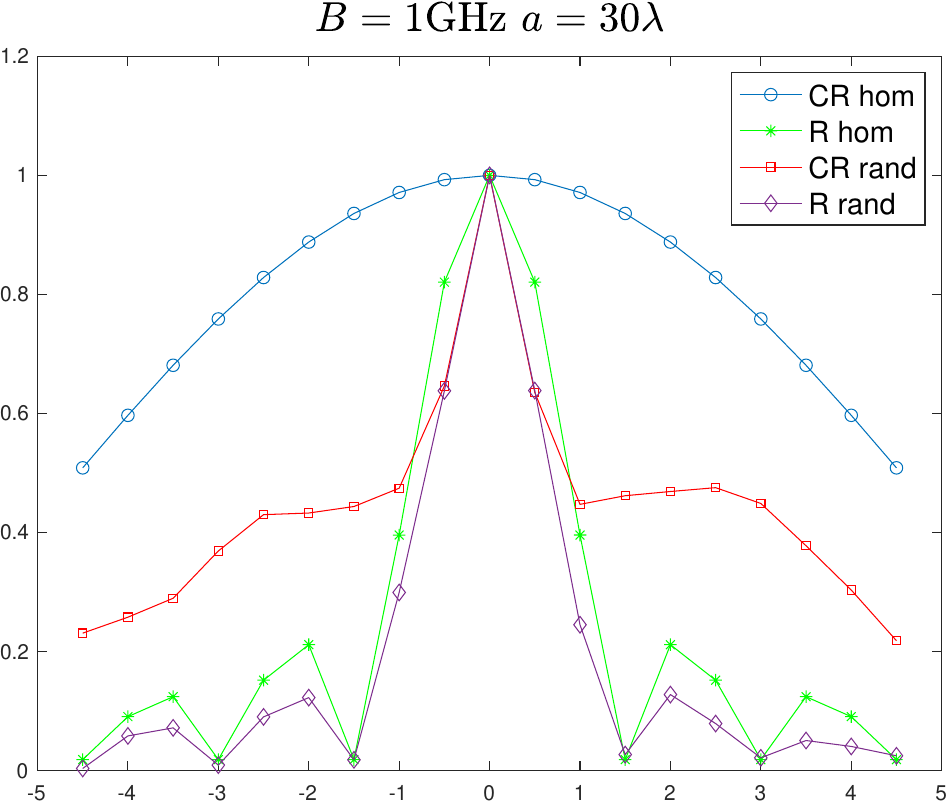}
   \includegraphics[width=0.31\linewidth]{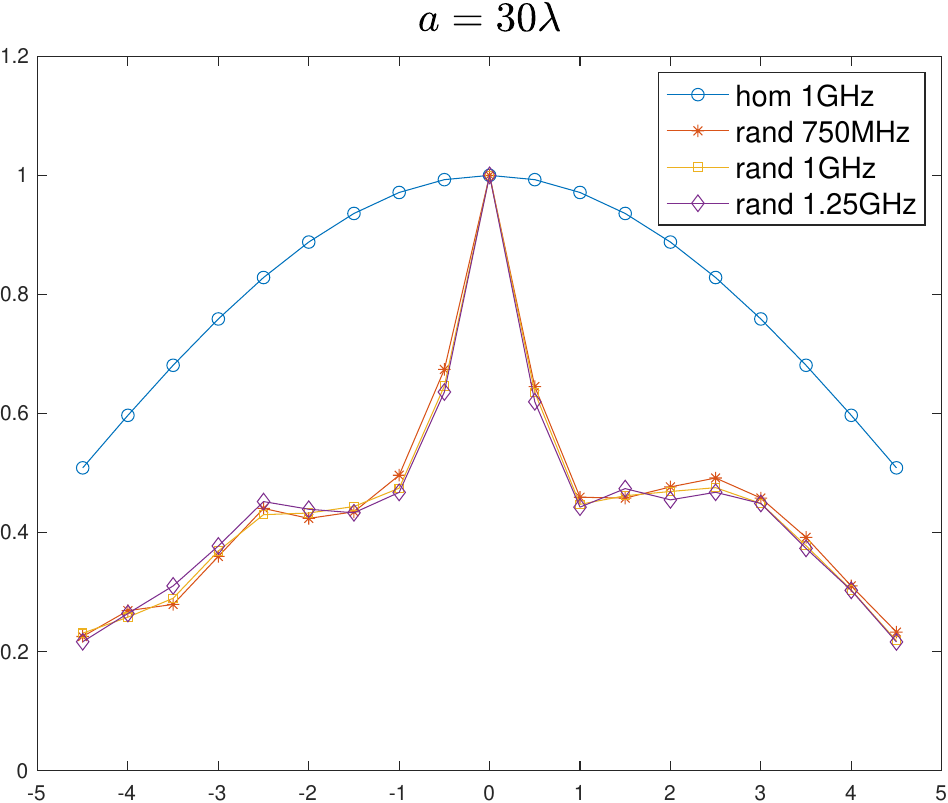}
    \includegraphics[width=0.31\linewidth]{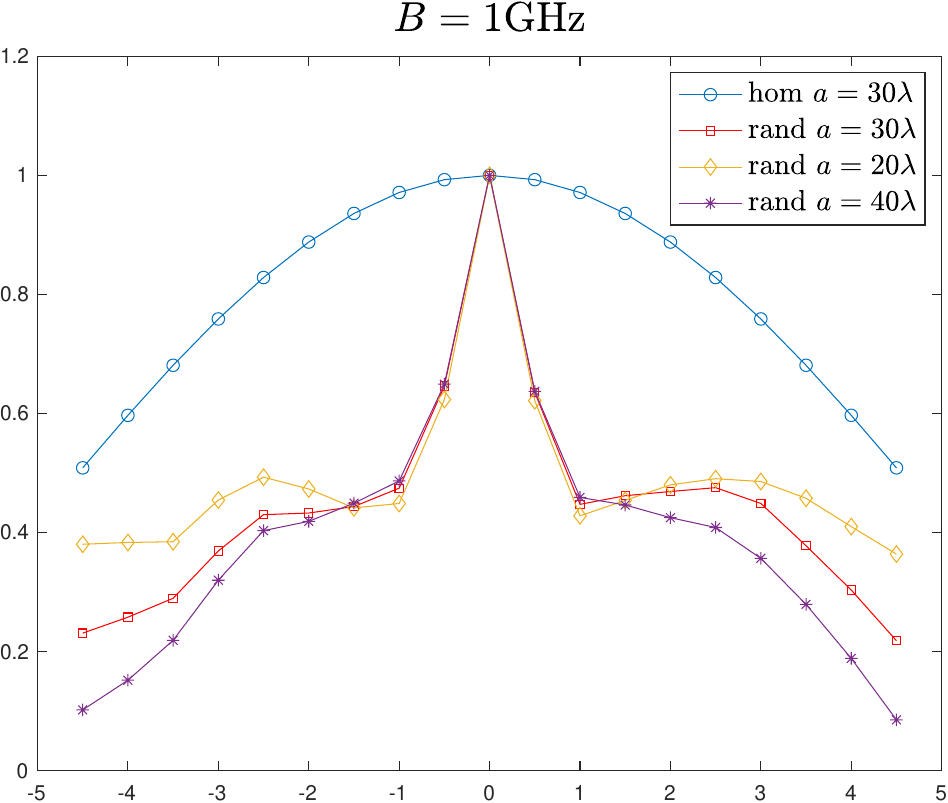}
\end{minipage}
\end{center}    
\caption{\label{fig:trFocus}  \textbf{Physical and effective aperture} Left:  Comparison of range and cross-range between random and homogeneous medium, with fixed array size and bandwidth. Super-resolution in cross-range is observed compared to the homogeneous medium, where resolution in cross-range is $\lambda L/a = 8\lambda$ and in range it is $c/B=5 \lambda$. 
  Middle: Effect of bandwidth with fixed array size: No effect on resolution, but a large enough bandwidth guarantees statistical stability over different random media with similar statistics. Right: Effect of physical array size with fixed bandwidth. The array size does not affect the main peak in different random media but the noise level off the main peak decreases with array size.}
\end{figure}

\begin{figure}[h!]
\begin{center}
\begin{minipage}{\linewidth}
  \includegraphics[width=0.3\linewidth]{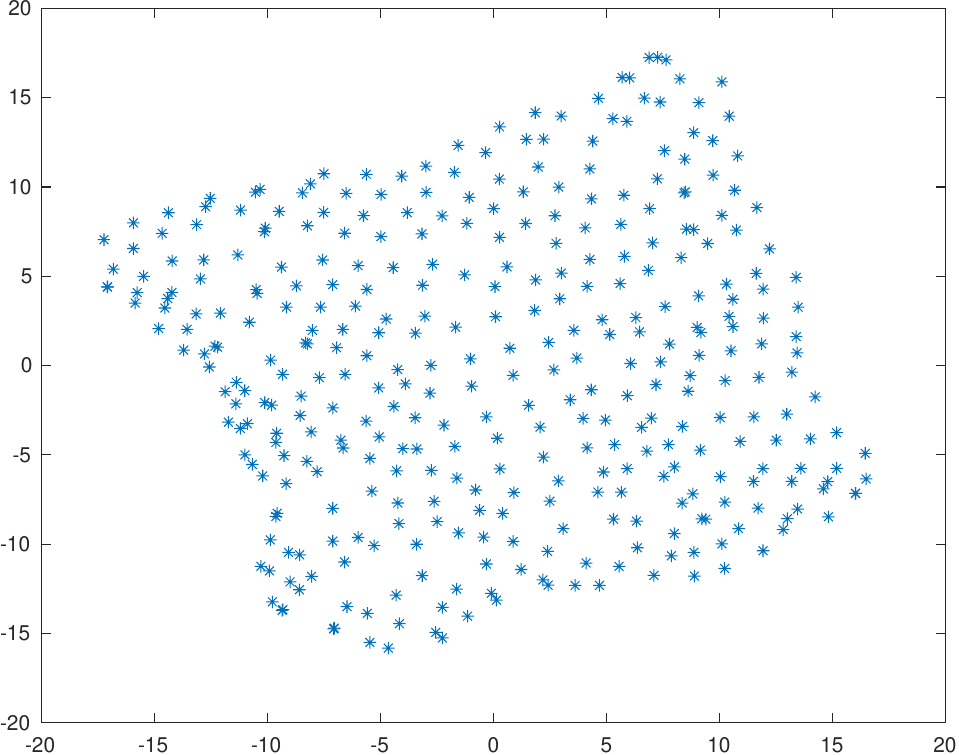}
  \includegraphics[width=0.3\linewidth]{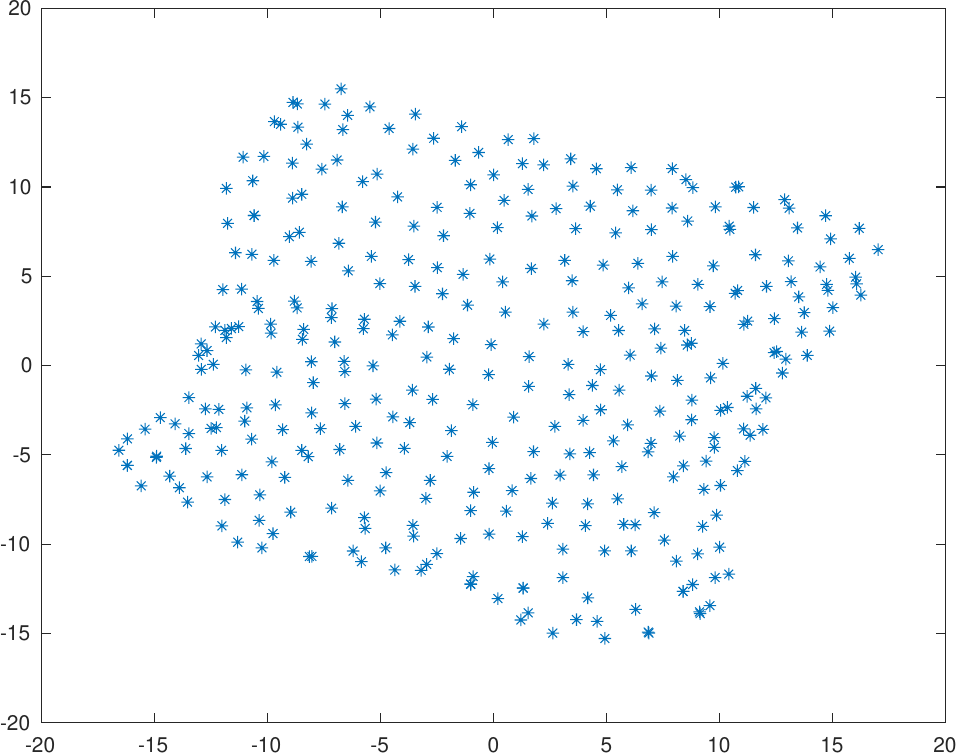}
    \includegraphics[width=0.3\linewidth]{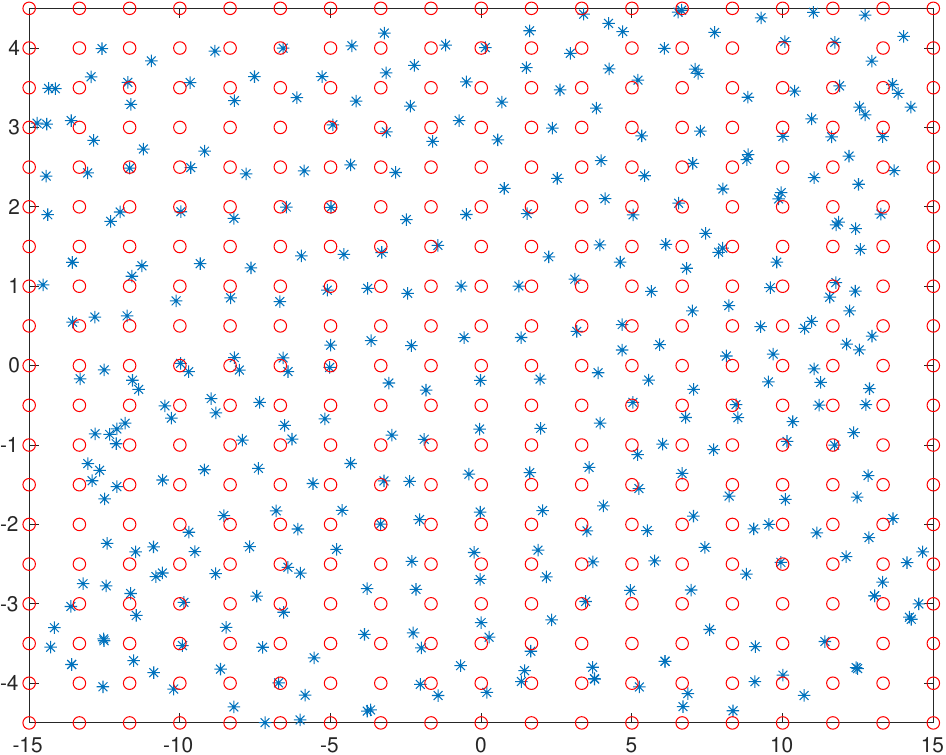}
\end{minipage}
\end{center}
\caption{\label{fig:grid} \textbf{Grid reconstruction using multidimensional scaling}. Left: the grid as reconstructed using the true Green's functions; Middle:  grid obtained from the estimated Green's functions using MDS; Right: the grid obtained from MDS after rotation and scaling assuming the location of 3 ``anchor" points is known, superimposed over true grid, plotted with red circles. This is the reconstruction that leads to the images shown in Figure \ref{fig:image}.}
\end{figure}

Figure~\ref{fig:trFocus} illustrates the super-resolution of time-reversal for the Foldy-Lax equations in our experimental setup. We fix the Green's function vector to a given source and compute its correlation with other Green's function vectors, while varying either the range coordinate or the cross-range coordinate of the latter.  We plot in Figure~\ref{fig:trFocus} the absolute value of this correlation to investigate the effect of random media on these
cross-correlations and hence the image formation. In the left panel of  Figure~\ref{fig:trFocus} we demonstrate the effect of randomness in the range and cross-range directions. 
The range resolution is only slightly better in the random medium and the side-lobes are somewhat lower than in the homogeneous medium. The cross-range resolution, however, significantly improves, as can be seen in the left panel of Figure~\ref{fig:trFocus}. Note also that the ordinate in the left panel takes values in $[-15 \lambda,15\lambda]$ while in the middle and left panels in $[-5 \lambda,5\lambda]$. In the middle panel of  Figure~\ref{fig:trFocus} we investigate how the bandwidth may affect super-resolution in the cross-range.  We observe that the effect is negligible. The bandwidth, however, plays a role in statistical stability, which can also be interpreted as a signal-to-noise ratio. 
In the right panel of Figure~\ref{fig:trFocus}, we investigate how the size of the array may affect super-resolution in cross-range.
We observe that the effect in resolution is negligible when the physical aperture varies at $10-40 \lambda$ (60-240cm). This stems from the fact that the effective aperture of the array is primarily determined by the characteristics of the random medium.  However, a wider physical receiver array leads to a reduction in the noise level outside the main peak. It is important to also note that the size of the array influences the amount of power it receives.

\begin{figure}[htbp]
\begin{center}
\begin{minipage}{9cm}
\hspace*{-0.1cm}
\includegraphics[width=4.35cm]{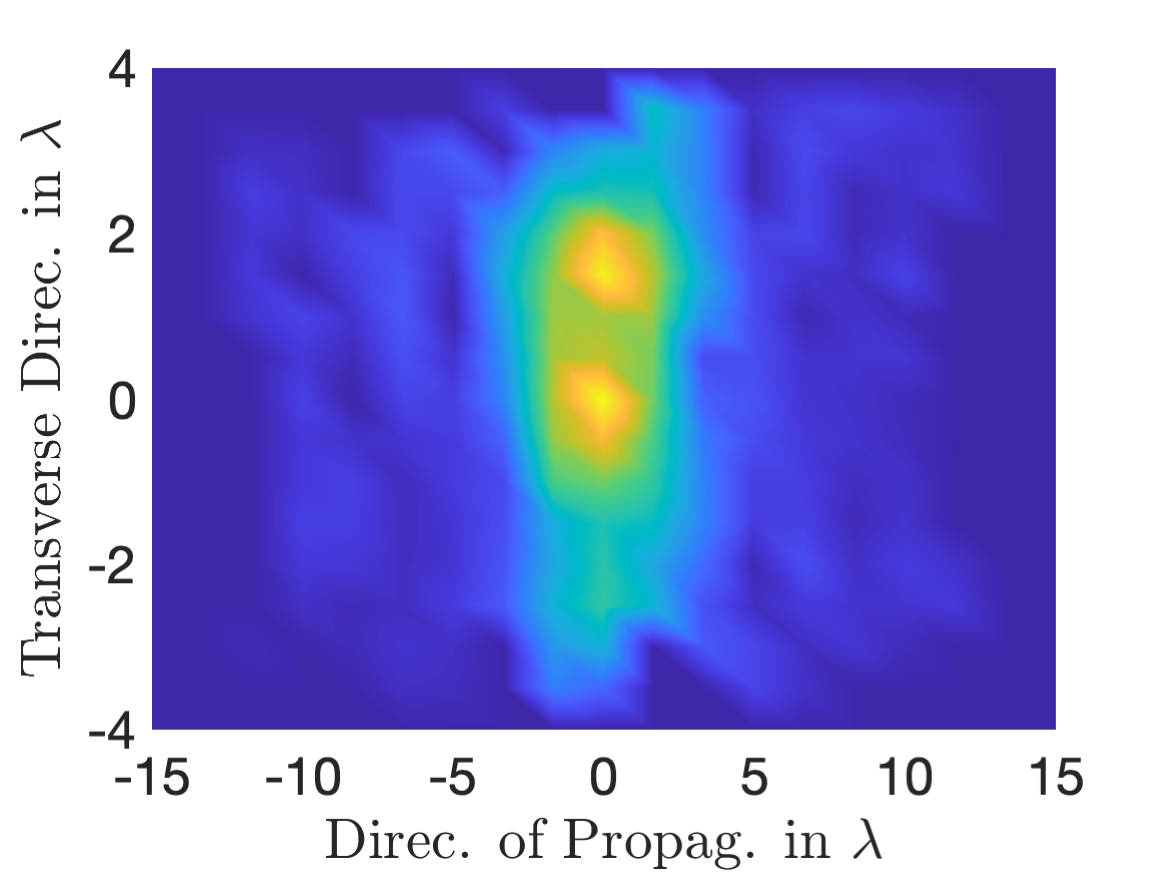} \hspace*{-0.35cm}
  \includegraphics[width=4cm]{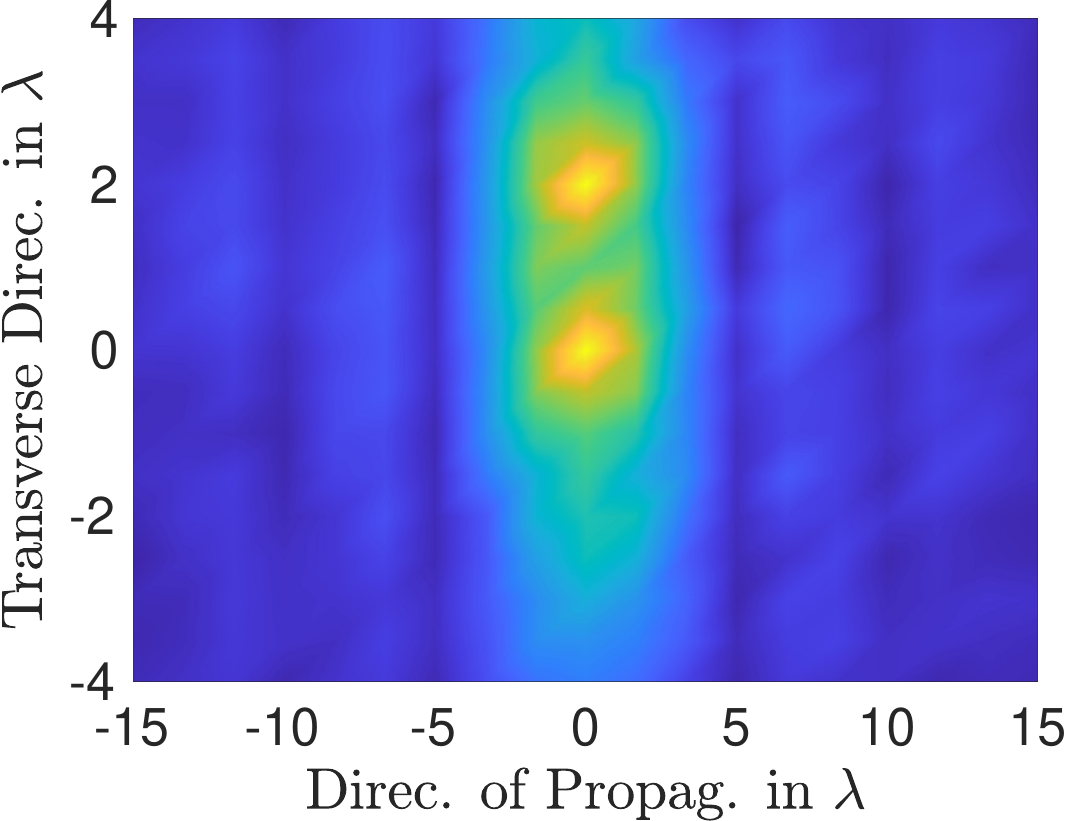}\\
    \includegraphics[width=4cm]{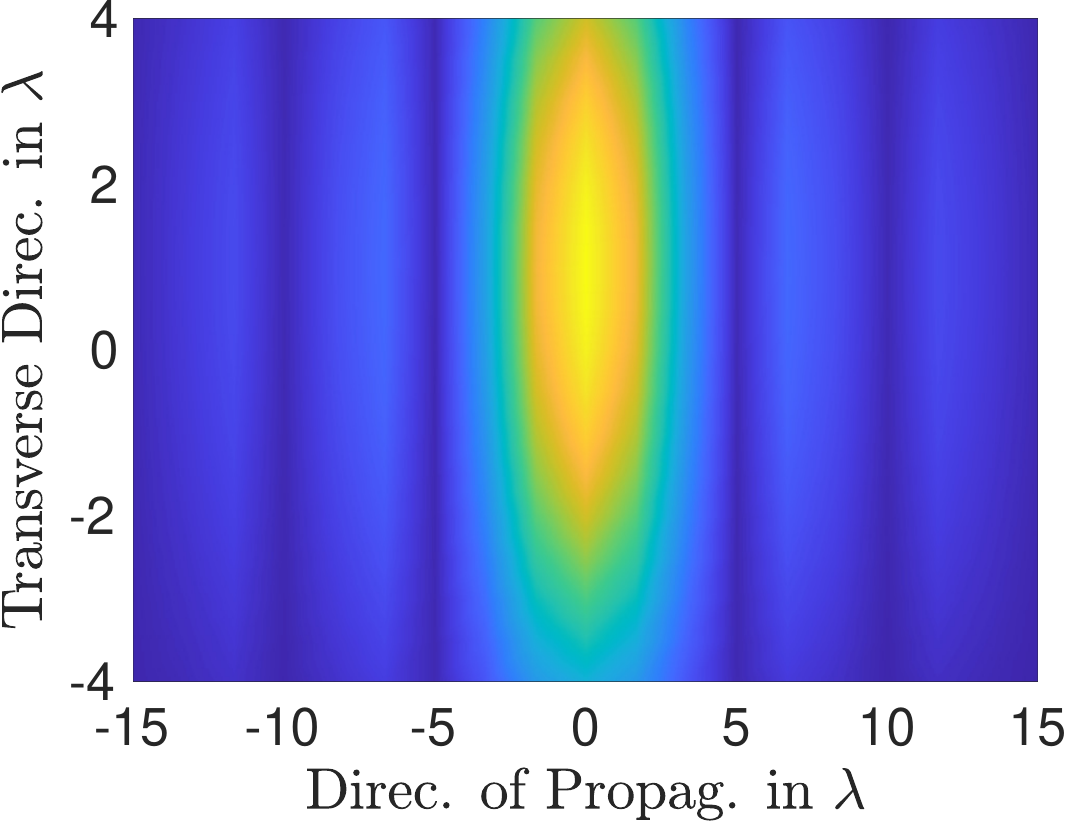}
  \includegraphics[width=4cm]{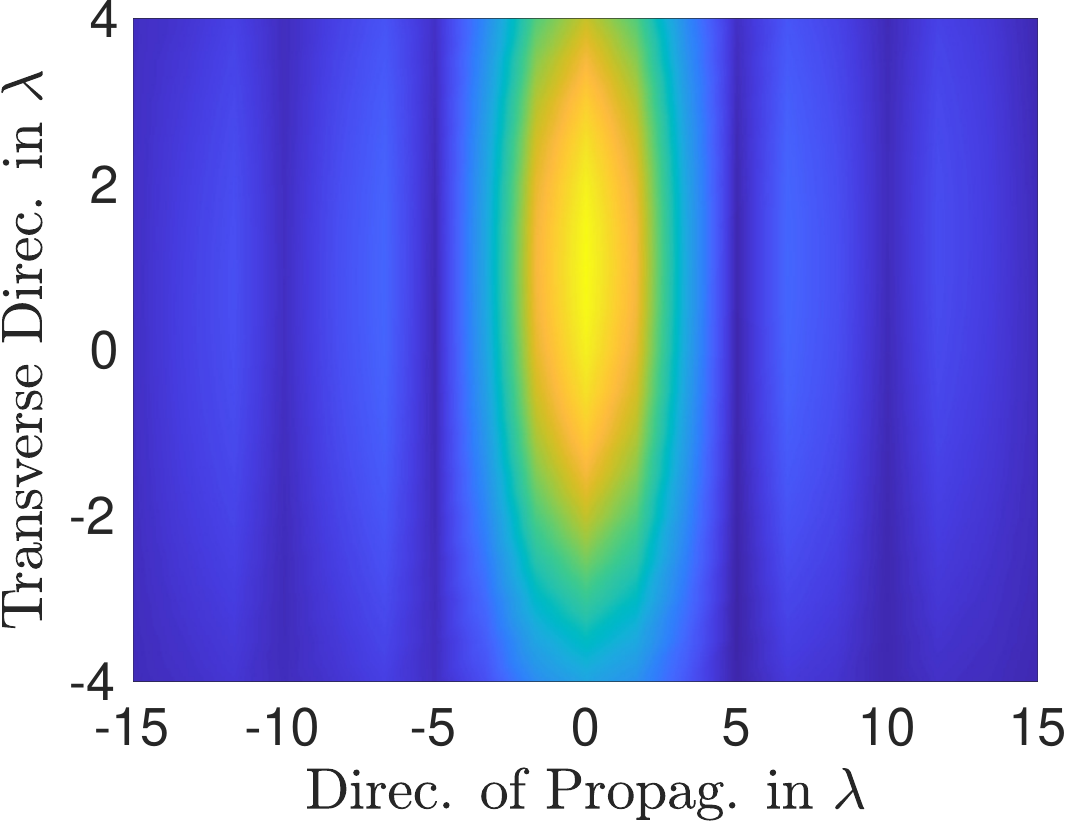}
\end{minipage}
\end{center}
\caption{\label{fig:image} \textbf{Super-resolution vs. time reversal}. {\bf Top left}: physical time reversal focusing on the known grid using the true Green's functions. {\bf Top right}: the proposed imaging algorithm, migration image using the estimated dictionary elements on the reconstructed grid from the ordered sensing matrix. We can see that we achieve similar resolution in both cases, which is able to resolve to closely located sources. {\bf Bottom left}: migration image in a homogeneous medium. {\bf Bottom right}: Homogeneous medium Green's function migration applied to random medium data. The resolution is inferior to that obtained by the estimated sensing matrix. Array size is $30 \lambda$ and the bandwidth is $1GHz$, with a $5GHz$ central frequency. }
\end{figure}

Figure~\ref{fig:grid} shows the outcome of the grid reconstruction step. The quality of this reconstruction is a key contribution of this paper.  An important fact is that we need the columns of the sensing matrix 
to be relatively incoherent for the dictionary learning steps to work, whereas they need to be relatively coherent in order to determine neighbors in the third step. 
In \cite{MNPT_PNAS}, we overcame this issue in weakly scattering media by decreasing the size of the physical array for the third step. This decreases resolution and allows for more accurate determination of neighbors. Here, we overcome this by imposing physical array sizes and bandwidths large enough to ensure coherence conditions are met.

As illustrated in the right panel of Figure~\ref{fig:trFocus}, changing the physical array size does not noticeably affect the cross-range resolution in strongly scattering media. This makes grid reconstructions in strongly scattering media more challenging because one cannot use subarray data to adjust the connectivity needed in the third step was done in~\cite{MNPT_PNAS}. Nevertheless, for a sufficiently broad bandwidth of 1GHz or more,
and for an array of about $20 \lambda$, we are able to recover the 4 nearest neighbors (2 from the cross-range and 2 from the range) in a stable manner for each pixel required for the MDS algorithm. As noted, a wider bandwidth and a larger physical array lead to a more robust neighborhood identification in our experiments.

Figure~\ref{fig:image} illustrates the migrated image
\begin{equation}
\label{eq:imag} 
{\mathcal I}(\vx_i)=  \left|  {\bfv}(\vx_i)^* \, \byy \right|\, , \quad i=1,\ldots,K, 
\end{equation} 
formed at points $\vx_i$ in the image window using the vector ${\bfv}(\vx_i)$ to back-propagate the recorded array data $\byy$.  In the figure, the recorded array data $\byy$ come from two nearby sources in the image window. In the top left image 
${\bfv}(\vx_i)={\vect\wg}(\vx_i)$, so we back-propagate $\byy$ using the true Green's function vectors ${ \vect\wg}(\vx_i)$. This is time reversal in the true random medium that shows the minimum refocusing spot sizes that can be achieved at the sources. In the top right image ${\bfv}(\vx_i)=\hat{\vect\wg}(\vx_i)$, the Green's function vectors estimated by our algorithm. Because we were able to obtain an accurate grid reconstruction, 
we achieve high-resolution images with the same resolution as time-reversal. 
In particular,  super-resolution allows us to image both targets, whereas standard Kirchhoff Migration with ${\bfv}(\vx_i)={\vect\wg}_0(\vx_i)$ results in the targets being blurred into one spot, as it is shown in the bottom right image. 

For comparison purposes, we also show in the bottom left image of Figure~\ref{fig:image} the result of  Kirchhoff Migration with ${\bfv}(\vx_i)={\vect\wg}_0(\vx_i)$ when we record the data $\byy$ in a homogeneous medium. This shows the diffraction limited resolution that is proportional to $c_0/B$ in range and to $\lambda L/a$ in cross-range.

\section{Unlabeled dictionary learning for imaging with neural networks}
\label{sec:NN}
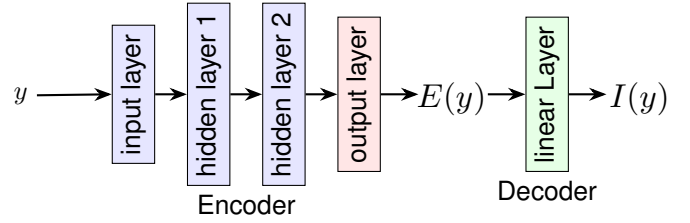
\begin{figure}[ht]
    \centering
    \begin{tikzpicture}[
        font=\sffamily,
        >={Stealth[bend]},
        block/.style={draw, rectangle, minimum height=1.6em, minimum width=5em, rotate=90,},
        decide/.code={%
            \ifnum#1=1
                \tikzset{fill=blue!10, execute at begin node={input layer}}
            \else\ifnum#1=2
                \tikzset{fill=blue!10, execute at begin node={hidden layer 1}}
            \else\ifnum#1=3
                \tikzset{fill=blue!10, execute at begin node={hidden layer 2}}
            \else\ifnum#1=4
                \tikzset{fill=red!10, execute at begin node={output layer}}
            \else\ifnum#1=5
                \tikzset{fill=green!10, execute at begin node={linear Layer}}
            \fi\fi\fi\fi
        }
    ]

        \begin{scope}[start chain=A placed {at={(\tikzchaincount*1-2,0)}},  
            nodes={on chain, block, join= by {thick,->}, decide=\tikzchaincount}]
            \path foreach \X in {1,...,4}{node{}};
        \end{scope}
        
        \begin{scope}[start chain=C placed {at={(\tikzchaincount*1+3.5,0)}},  
            nodes={on chain, block, join= by {thick,->}, decide=5}]
            \path foreach \X in {5}{node{}};
        \end{scope}
        
        \node[anchor=south] (input) at (-2.5,-.25) {$y$};  
        \draw[thick,<-] (A-1.north) -- (input);
        
        \draw[thick, ->] (A-4.south) -- ++ (.5, 0) node [scale=1.25, anchor=south,shift={(1em,-1em)}, rotate=0](output1){$E(y)$};
        \draw[thick, ->] (C-1.south) -- ++ (.5, 0) node [scale=1.25, anchor=south,shift={(1em,-1em)}, rotate=0](output3){$I(y)$};
        \draw[thick, <-] (C-1.north) -- ++ (-.5, 0) node [scale=1.25, anchor=south,shift={(0.5em,-0.5em)}, rotate=0](input3){};

        \node[shift={(.5,-.25)}]  at (A-2.west) {Encoder};
        \node[shift={(0,-.25)}]  at (C-1.west) {Decoder};

    \end{tikzpicture}
    \caption{\textbf{Neural network architecture.} We use a two hidden layer network for the encoder. The output of the encoder is fed into the linear decoder network. Modulus thresholding activation~\eqref{e:leakyrelu} is used in the red layers and LeakyReLU activation is used for the blue layers. Before activation functions are applied, the input is standardized using batch normalization, as described in~\cite{Ioffe2015}. The green layer is linear, without an activation function applied. The hidden dimensions of the encoder networks are $4096$, and $2048$, respectively. The decoder matrix has $1024$ columns. We train the networks to minimize the loss function $\mathcal{L}= \|I(y)-y\|^2_2+\mu \|\ E(y)\|_1.$}
    \label{f:arch}
\end{figure}

In this section, we describe how to recover the sensing matrix using multiple realizations of an encoder-decoder neural network, as illustrated in Figure~\ref{f:arch}. Each realization is obtained from training two networks to learn the forward and inverse problems described by
\begin{equation}
\label{e:axeqb}
\byy = \vect{\Gc}\, \bxx \, ,
\end{equation}
%
i.e., relating the sources $\bxx$ to the data $\byy$ and vice versa. 
The networks are trained and validated on data sets in $\C^{806}$ consisting of $5000$ and $3000$ data points respectively. 

Both sets are generated with the Foldy-Lax model~\ref{ss:FL} as in Section \ref{s:numersim},  whose specification we repeat here for completeness of the neural network section. The imaging system operates with a central frequency of $f_0=5$GHz, wavelength $\lambda=6cm$,  and bandwidth $B=1$GHz discretized into $26$ evenly spaced frequencies. The imaging window dimensions are defined with a cross-range spacing of $dl_x = 0.5 \lambda$ and a range spacing of $dl_z=1.66 \lambda$. This produces a total of $K=361$ pixels ($nx=nz=19$) in the image window. The scattering area is  is taken to be 10$\times$10 meters and is populated with  $400$ randomly distributed scatterers. Lastly, an array consisting of  $30$ transducers with distance $\lambda$ between them is positioned at a distance of 14 meters from the center of the imaging window. Individual data points in the training and validation sets are generated by randomly positioning three point sources (sparsity $s=3$) in the image window.

The two networks recover an unordered collection of vectors containing estimates of the columns of $\vect{\Gc}$. The estimation follows two steps: In {\it Step One}, the Green's function vector estimates are obtained from the weight matrix of the linear decoder network. A single realization of the encoder-decoder network does not recover accurately all of the columns of the sensing matrix. However, different initializations of the neural network recover different columns with sufficient accuracy. We thus recover a large set, containing many versions of the columns of $\vect{\Gc}$ by training multiple encoder-decoder network realizations. This large set contains, in particular, multiple accurate versions of the columns of $\vect{\Gc}$, in addition to extraneous vectors, to be discarded.

In {\it Step Two} we use the Density-Based Spatial Clustering of Applications with Noise (DBSCAN) algorithm of \cite{Ester1996}  to cluster the large collection of columns of $\vect{\Gc}$. Taking a representative member or an average of the columns of each cluster produces an accurate, unordered, dictionary for the 
columns of $\vect{\Gc}$.

Following this step the MDS method of Section \ref{sec:MDS} is used again to order the columns.

\subsection{The Encoder-Decoder}\label{s:ed}
We wish to represent to forward problem of~\eqref{e:axeqb} ($\bxx\rightarrow \byy$) using a linear network $D$ and the inverse problem ($\byy\rightarrow \bxx$) using 
a fully connected network $E$, as shown in  Figure~\ref{f:arch}. By fully connected, we mean a network containing $k$ hidden layers, where the $i$-th layer is represented by a complex affine transformation dimensions $d_{i-1}\to d_i$, followed by a nonlinear activation function $\alpha_i$.
\begin{equation}\label{e:fcet}
N_{F}(z)=L_{out}(h_i(h_{i-1}(\dots h_1(h_0(z))\dots))), i=0,\dots k,
\end{equation}
and
\[
h_i=\alpha_i\circ L_i, \hspace{0.2em} L_i(x)= W_ix+b_i,\hspace{0.2em} W_i\in \C^{d_{i}\times{d_{i-1}}},b_i\in \C^{d_{i}}
\]
We use either Leaky ReLU $ReLU_\epsilon$ or modulus thresholding $T_\epsilon$ as nonlinear activation functions. LeakyReLU is evaluated on the imaginary and real components separately. The two activation functions can be written as 
\begin{equation}\label{e:leakyrelu} 
\begin{split}
ReLU_\epsilon(z)=&\max(0, real(z))+\epsilon \min(0, real(z)) + \\
&i(\max(0, imag(z))+\epsilon \min(0, imag(z))
\end{split}
\end{equation}
and
\begin{equation}\label{e:modt}
T_\epsilon(z)=\frac{z \max(0, |z|-\epsilon)}{|z|}.
\end{equation}
We use modulus thresholding in the image space (last layer) and LeakyReLU otherwise, i.e 
\begin{align}
&\alpha_i=ReLU_\epsilon, i=0,\dots,k-1\\
&\alpha_k =T_\epsilon\nonumber
\end{align}

The architecture used for the numerical experiments is shown in Figure~\ref{f:arch}. We use a two-layer network for the encoder and a linear network with $K_{net}$ columns for the decoder. The $K_{net}$ columns are interpreted as Green's vector estimates, that is, columns of $\vect{\Gc}$. The number of layers in the encoder and the number of columns in the decoder weight matrix are somewhat arbitrary but larger than what is expected. This is because the clustering step, following the network training, prunes the columns and selects an accurate collection of the columns of $\vect{\Gc}$. Thus, by training multiple encoder-decoder networks and clustering column estimates it is unnecessary to try and recover the correct number of Green's function vectors in any single realization. In preliminary testing, we found that using $K_{net}>K$ required significantly fewer encoder-decoder realizations to achieve accurate final estimates (see Figure~\ref{f:knet}). Because of this, we use $K_{net}=1024$ in our numerical experiments instead of $K_{net}=K=361$.

\subsection{Loss function}\label{s:loss}
The loss function used for training contains two terms. Recall that we define $E$, the encoder,  and $I=D\circ E$, the composition network. We apply mean square error loss to the output of the $I$ compared with the input. On the encoder output we apply a sparsity promoting $L^1$ loss, as we expect source configurations to be sparse. With a tunable hyperparameter $\mu$, 
the loss is 
\begin{equation}\label{e:unlabloss}
   \mathcal{L}=  \|I(y)-y\|^2_2+\mu \|E(y)\|_1.
\end{equation}
We minimize~\eqref{e:unlabloss} for the encoder and decoder simultaneously. If the weights are instead updated iteratively (separately for the encoder and decoder), then a similar performance is observed.

\subsection{Network hyperparameters}\label{s:hyper}
The numerical experiments were coded using PyTorch.  AdamW  optimizer \cite{hutter2018} was used to update the network weights. We use default hyperparameters apart from the learning rate which is taken to be $10^{-5}$. We apply batch normalization before each activation function as described in~\cite{Ioffe2015}. The values for $\epsilon$ for $ReLU_\epsilon,T_\epsilon$ are both taken to be $.001$. The $L^1$ weight, $\mu$, in~\eqref{e:unlabloss} is taken to be $.1$. All networks are trained for $10000$ epochs. Learning stagnates after around $5000$ epochs.

\begin{figure}
\centering
\begin{tikzpicture}
\pgfmathsetmacro{\xmin}{0}
\pgfmathsetmacro{\xmax}{.75}
\pgfmathsetmacro{\ymin}{0}
\pgfmathsetmacro{\ymax}{.75}
\begin{groupplot}[
  group style={
    group name=my plots,
    group size=2 by 2, 
    vertical sep=0.01cm,
    horizontal sep=0.01cm
  },
  width=6cm, 
  hide axis,
  enlargelimits=false,
  axis equal image,
]
  \nextgroupplot
    \addplot graphics [xmin=\xmin,xmax=\xmax,ymin=\ymin,ymax=\ymax]
          {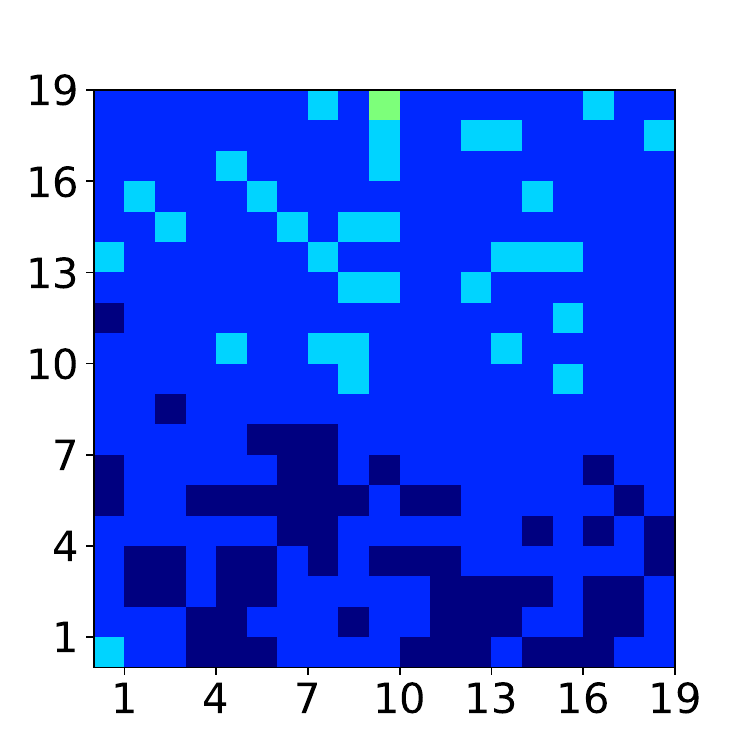};
    \coordinate (top) at (rel axis cs:0,1);
  \nextgroupplot
    \addplot graphics [xmin=\xmin,xmax=\xmax,ymin=\ymin,ymax=\ymax]
          {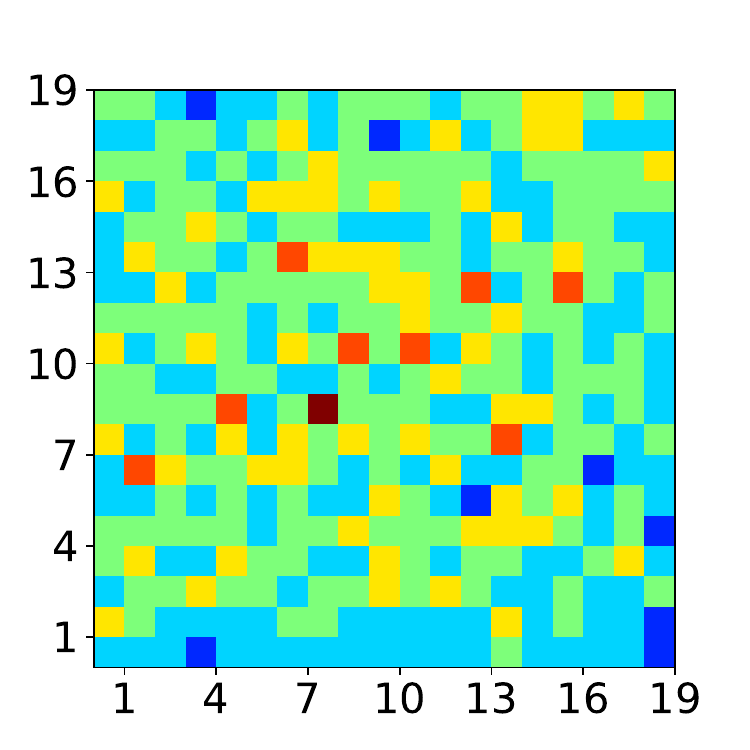};
  \nextgroupplot
    \addplot graphics [xmin=\xmin,xmax=\xmax,ymin=\ymin,ymax=\ymax]
          {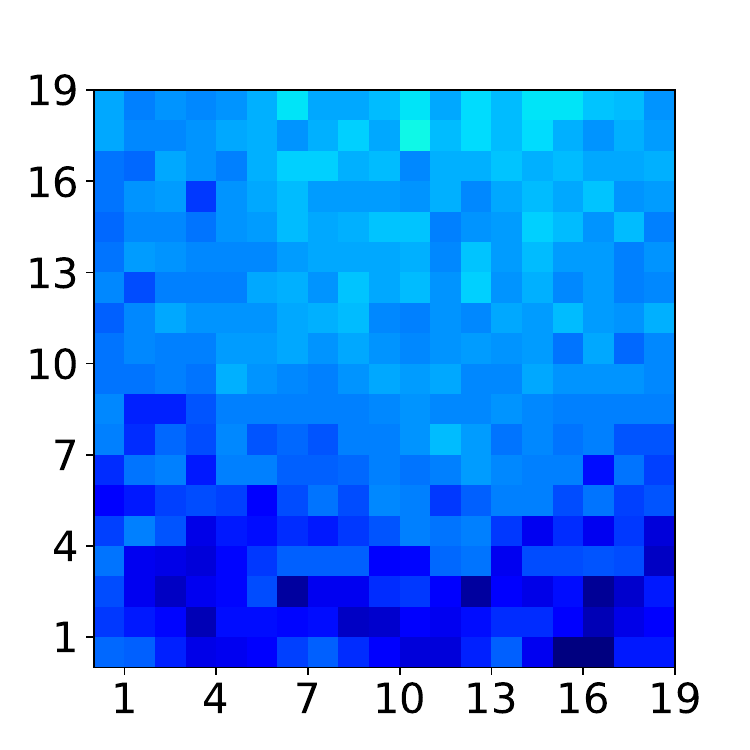};
  \nextgroupplot
    \addplot graphics [xmin=\xmin,xmax=\xmax,ymin=\ymin,ymax=\ymax]
          {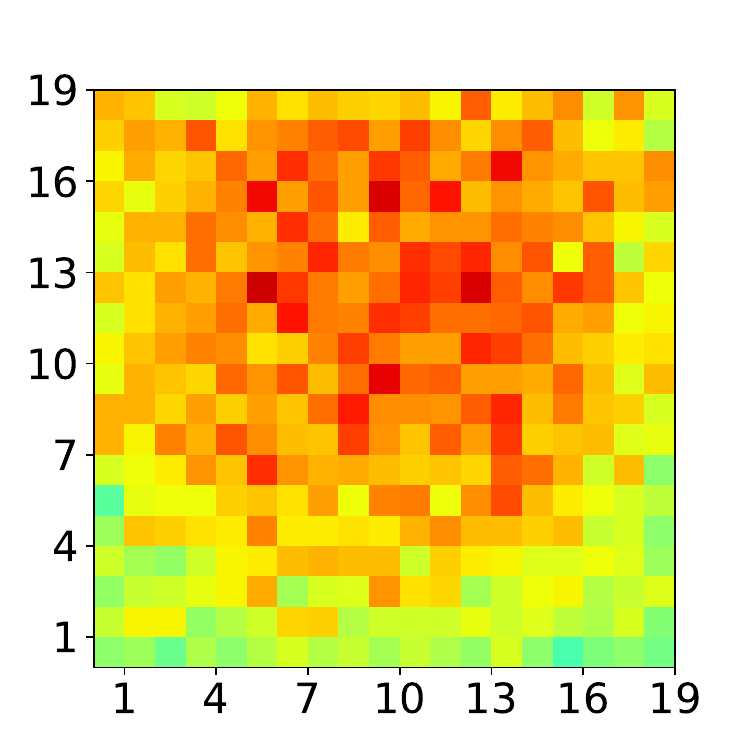};
    \coordinate (bot) at (rel axis cs:1,0);
\end{groupplot}

\path (top-|current bounding box.east) coordinate (legendposright_top);
\path (bot-|current bounding box.east) coordinate (legendposright_bottom);

\begin{axis}[%
  hide axis,
  scale only axis,
  height=0.45\linewidth,
  at={(legendposright_top.east)},
  anchor=east,
  xshift=0.15cm,
  yshift=-3.2cm,
  point meta min=0.0,
  point meta max=6.0,
  colormap/bluered,
  colorbar right,
  colorbar sampled,
  colorbar style={samples=256},
  every colorbar/.append style={
    height=.5*\pgfkeysvalueof{/pgfplots/parent axis height},
    width=.03*\pgfkeysvalueof{/pgfplots/parent axis width}
  }
]
  \addplot [draw=none] coordinates {(0,0)};
\end{axis}

\begin{axis}[%
  hide axis,
  scale only axis,
  height=0.45\linewidth,
  at={(legendposright_bottom.east)},
  anchor=east,
  xshift=0.15cm,
  yshift=.37cm,
  point meta min=0.0,
  point meta max=100.0,
  colormap/bluered,
  colorbar right,
  colorbar sampled,
  colorbar style={samples=256},
  every colorbar/.append style={
    height=.5*\pgfkeysvalueof{/pgfplots/parent axis height},
    width=.03*\pgfkeysvalueof{/pgfplots/parent axis width}
  }
]
  \addplot [draw=none] coordinates {(0,0)};
\end{axis}
\end{tikzpicture}
\vspace{-40pt} 
\caption{\textbf{Using $K_{net}>K$ requires less realizations to recover all columns of the sensing matrix.} We use the true sensing matrix to count the number of times each Green's vector is recovered with $99\%$ accuracy. Each subfigure shows the image window with the pixel value being the number of times the corresponding Green's vector is recovered. The colorbar shows the scale of each row. {\bf Top left}: $1$ network realization with $K_{net}=K=361$ decoder columns. Here $61$ out of $361$ columns are not recovered at all, and $1$ column is recovered $3$ times. {\bf Top right}: $1$ network realization with $K_{net}=K=1024$ decoder columns. All columns are recovered at least once, and $353$ out of $361$ are recovered more than once. {\bf Bottom left}: $25$ network realizations with $K_{net}=K=361$ decoder columns. Here $2$ columns are not recovered at all and $359$ out of $361$ columns are recovered at least twice. {\bf Bottom right}: $25$ network realizations with $K_{net}=1024$ decoder columns. All columns are recovered at least $44$ times. The maximum number of times a column is recovered is $93$. We train the networks to minimize~\eqref{e:unlabloss} with $\mu=.1$. Each network is trained with a different random initialization and a different seed used for gradient descent. The network architecture is as in Figure~\ref{f:arch}.}
\label{f:knet}
\end{figure}
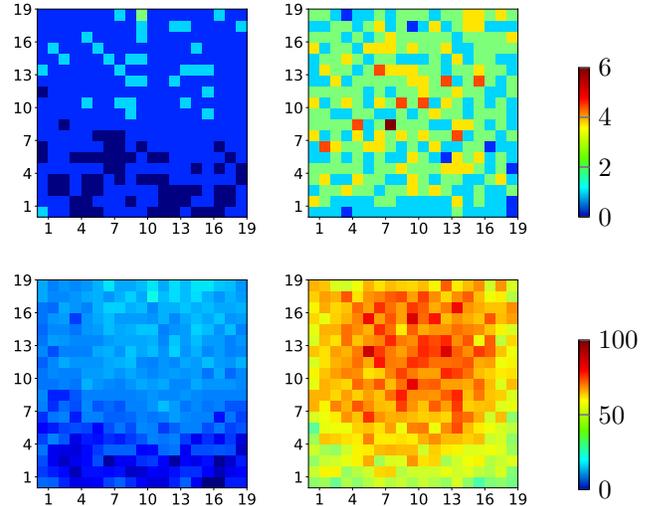

\subsection{Clustering using DBSCAN}
Training a single encoder-decoder network realization does not recover an accurate enough dictionary for the columns of $\vect{\Gc}$ so that the MDS algorithm of Section \ref{sec:MDS} can be applied. In order to get an accurate dictionary, we train $L$ different realizations of initialized weights. Training the $L$ networks produces $L$ realizations of our estimated, unordered, sensing matrix,
\[
\hat{\vect{\Gc}}^{(i)}=[\hat{g}^{(i)}_{1},\dots\hat{g}^{(i)}_{K}],~i=1, \dots~ L,
\]
obtained from the weight matrices of the decoder networks. An improved dictionary is obtained by clustering the columns $\hat{g}^{(i)}_{j}$ using the Density-Based Spatial Clustering of Applications with Noise (DBSCAN) algorithm of \cite{Ester1996}, and selecting the average of the oriented vectors from the $K$ largest clusters as the candidate $K$ columns of $\hat{G}$. The distance function used to cluster the $L^2$ normalized collection of vectors $\{\hat{g}^{(i)}_{j}\}$ is
\begin{equation}\label{e:dbsscandist}
d(z, \tilde{z})=1-|\left\langle z , \tilde{z}\right\rangle|,
\end{equation}
where all vectors are normalized unit vectors.
Since the strengths of the source configurations in the image window are unknown, the signs of the recovered columns may be incorrect. Using~\eqref{e:dbsscandist} as a metric addresses this issue since collinear vectors are considered to be the same. 
 
The DBSCAN algorithm forms and expands clusters about areas with high point density and labels points in low-density areas as noise. Given the  distance function $d(z,\tilde z)$ and parameters
\[
\varepsilon>0\text{ and } C_{min}\in\mathbb{Z},
\] DBSCAN returns a mapping from the data to the corresponding cluster. If two points are within distance $\epsilon$ then they are considered neighbors. If a point has $C_{min}$ neighbors it is considered a core sample in a cluster. Clusters are then expanded about core samples. 

The DBSCAN algorithm processes the data points as follows: Start with a data point $z$ that has not been assigned to a cluster. If there are at least $C_{min}$ points within distance $\varepsilon$ from $z$, form a new cluster $C$ consisting of the data points in the epsilon ball around $z$ $B_\varepsilon(z)$ that have not yet been clustered. For each point $\tilde{z}\in C$ with at least $C_{min}$ points within distance $\varepsilon$ from $\tilde{z}$, expand the cluster by including the elements of the epsilon ball $B(\tilde{z}, \epsilon)$ that have not been assigned a cluster. The algorithm terminates after all data points have been processed. That is, after all points are assigned to a cluster or assigned as noise.

The final estimate of the unordered sensing matrix is obtained by taking the average of the oriented vectors in the $K$ largest clusters found by DBSCAN. Since the Green's function vector estimates for the same focal point from different network realizations may differ by a sign, they must be oriented to point in the same direction before averaged. The Green's vector estimate from cluster $C$ is the average over the set
\begin{equation}\label{e:orientave}
\tilde{C}=\{z : z \text{ or } -z\in C \text{ and } \left\langle z, v \right \rangle > 0 \},
\end{equation}
where $v$ is some arbitrary fixed reference vector in the cluster. Using different reference vectors will only affect the sign of the estimate.

\begin{figure}[ht]
     \centering
         \begin{subfigure}[t]{0.35\textwidth}
         \centering
         \includegraphics[width=.80\textwidth]{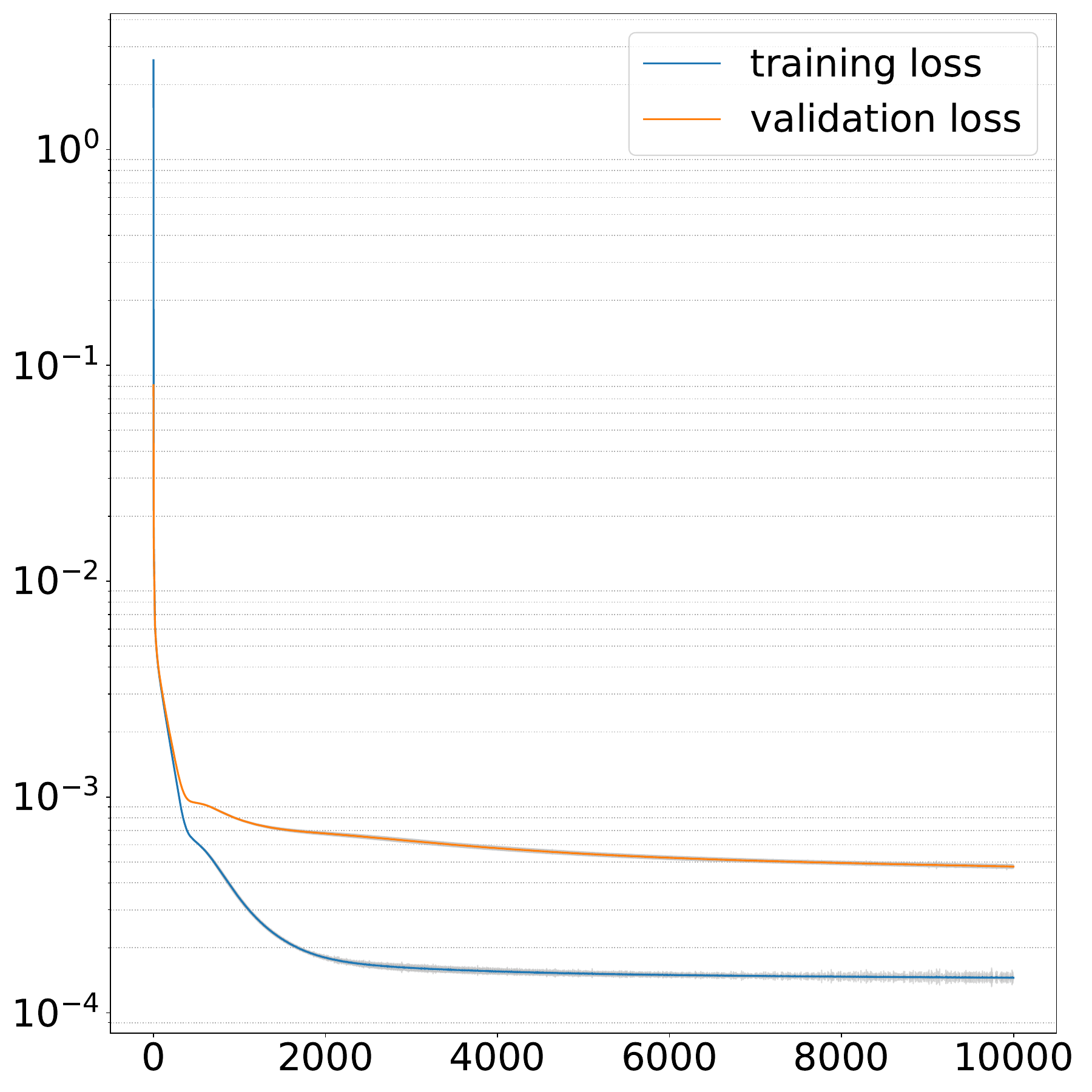}
     \end{subfigure}
     \vskip 0.5em
          \begin{subfigure}[t]{0.35\textwidth}
         \centering
          \includegraphics[width=.80\textwidth]{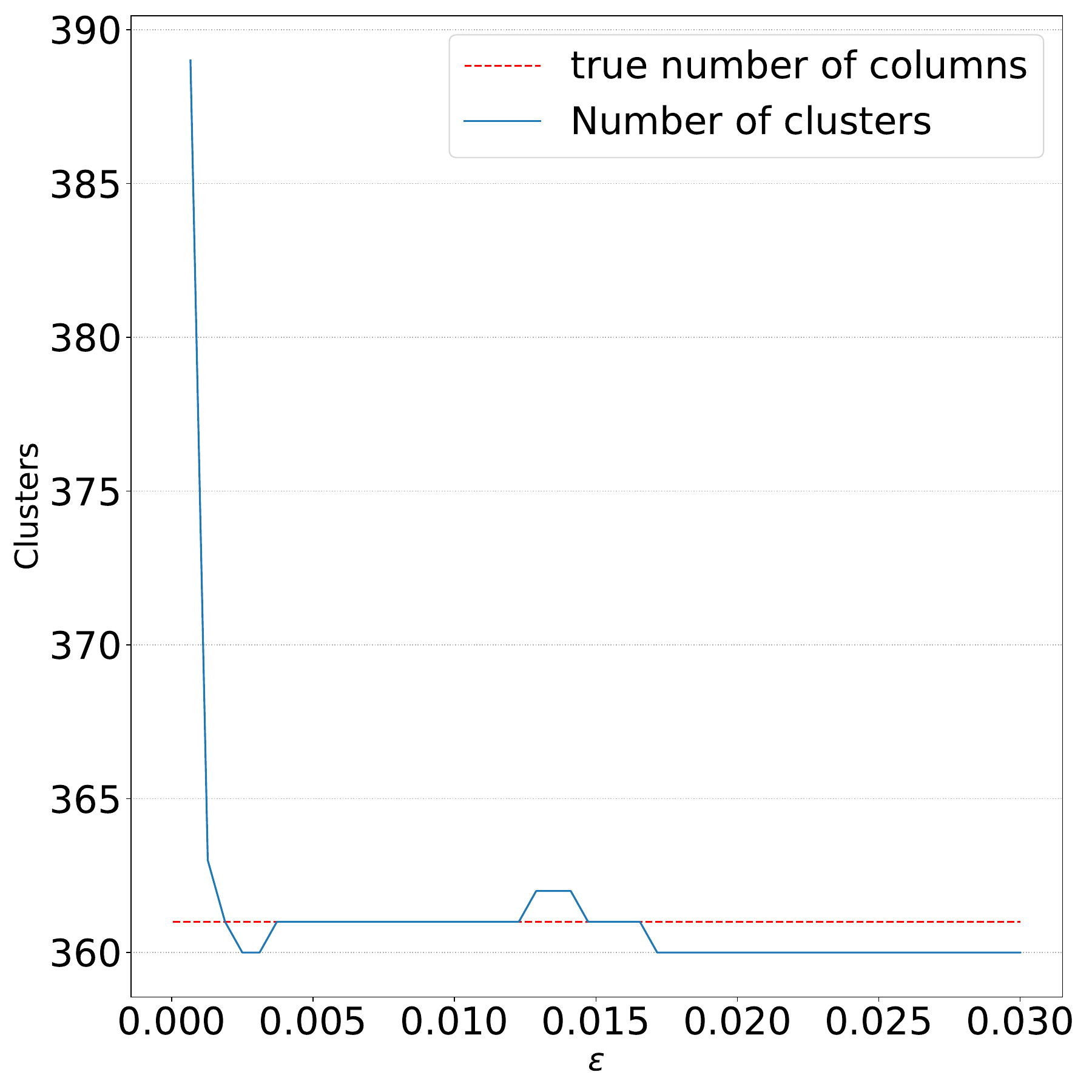}
     \end{subfigure}
\caption{\textbf{Clustering of encoder-decoder realizations} Top: Average training and validation loss curves across $25$ network realizations. Bottom: Number of clusters outputted by the DBSCAN algorithm with minimum neighbors $C_{min}=5$ vs. threshold hyperparameter $\varepsilon$. We train $L=25$ networks to minimize~\eqref{e:unlabloss} with $\mu=.1$. Each network is trained with a different random initialization and a different seed used for gradient descent. The network architecture is as in Figure~\ref{f:arch}. We see that the number of clusters found by DBSCAN is constant and equal to $K=361$, the true value, across a range of threshold values. If more than $K$ clusters are recovered, then one can work solely with the $K$ largest clusters or use a different value for the $C_{min}$ parameter. 
When extra noise clusters are found they contain significantly fewer vectors than clusters corresponding to Green's function vectors.} 
\label{f:lossclusterhisto}
\end{figure}

Given enough realizations, multiple estimates of each true Green's function vector will be present in the collection $\{\hat{g}^{(i)}_{ j}\}$ (see Figure~\ref{f:knet}~(d)). The parameter $\varepsilon$ can be fine tuned on a further test set if the original value does not yield a perfect sensing matrix. After clustering using DBSCAN with parameters $C_{min}=5$ and $\varepsilon$ in the range $0.005-0.01$ (see Figure~\ref{f:lossclusterhisto}), we see that $K=361$ emerges as a good estimate of the true number of points in the image window, when $K$ is not known in advance, at which point the averages of the oriented vectors in the clusters approximate all columns of the true sensing matrix with better than $99\%$ accuracy.

\section{Summary and conclusions} 
\label{sec:sumc}
We have introduced a computational imaging approach that allows for high resolution images in strongly scattering media. The proposed methodology leverages large and diverse data sets to estimate the sensing matrix of a strongly scattering random medium. Super-resolution is obtained by exploiting the fact that scattering effectively enhances the physical aperture. The stability of the imaging algorithm depends on the size of the physical array and on the bandwidth.

We use two different approaches to estimate the sensing matrix from large and diverse data sets, one that relies on conventional optimization approach (Section \ref{sec:3step}) and one that uses neural networks (Section \ref{sec:NN}). The  first method involves three steps: Getting a preliminary, low quality estimate of the columns of the sensing matrix, applying a non-convex optimization that improves its accuracy when initialized by the preliminary estimate, and, finally, using a neighborhood based (non-metric) multidimensional scaling algorithm to order the estimated columns of the sensing matrix, which is essential for imaging (Section \ref{sec:MDS}). 

In the neural network based approach, we use an encoder-decoder architecture trained on the diverse data, again without any knowledge of the source locations or of the properties of the random medium. We get accurate estimates of the unordered sensing matrix from multiple realizations of the trained network, followed by a clustering step using DBSCAN, (Section \ref{sec:NN}). This is followed by the same ordering step is the same of Section \ref{sec:MDS}. 

The main advantage of the neural network approach is that no initialization is needed. Accuracy is gained by generating multiple versions of the columns of the sensing matrix, followed by clustering. No elaborate fine-tuning of the hyper-parameters of the neural network is needed. The main advantage of the conventional non-convex optimization approach is that we have very good control of accuracy,  but a good initialization is necessary.

%
%

{\bf Acknowledgements.} Miguel Moscoso's work was  supported by  the Spanish AEI grant PID2020-115088RB-I00.
Alexei Novikov's work was partially supported by NSF 2407046, AFOSR FA9550-23-1-0352, and FA9550-23-1-0523. 
The work of Alexander Christie and George Papanicolaou was partially supported by AFOSR FA9550-23-1-0352.
The work of  Chrysoula Tsogka was partially supported by AFOSR FA9550-23-1-0352 and FA9550-24-1-0191.

\bibliographystyle{IEEEbib}

\begin{thebibliography}{90}
\bibitem{agarwal} A. Agarwal, A. Anandkumar, P. Jain, and P. Netrapalli, {Learning sparsely used overcomplete dictionaries via alternating minimization,} SIAM Journal on Optimization {\bf 26}, 2775--2799 (2016). 
\bibitem{BPTB02}  L. Borcea, C. Tsogka, G. Papanicolaou and J. Berryman, Imaging and time reversal in random media, Inverse Problems, {\bf 18}, 1247--1279 (2002). 
\bibitem{BPT06} 
L. Borcea, G. Papanicolaou and C.T.,  Adaptive interferometric imaging in clutter and optimal illumination, Inverse Problems, {\bf 22}, 1405--1436 (2006). 
\bibitem{BGPT11}
L. Borcea, J. Garnier, G. Papanicolaou and C. Tsogka, Enhanced statistical stability in coherent interferometric imaging, Inverse Problems, {\bf 27}, p. 085003 (2011). 
\bibitem{Fink2000} M. Fink, D. Cassereau, A. Derode, C. Prada, P. Roux, M. Tanter,
J.-L. Thomas, and F. Wu, Time-reversed acoustics, Rep. Prog. Phys. 63, 1933-1995 (2000)


\bibitem{Oh10} S. Oh, A. Montanari and A. Karbasi, {Sensor network localization from local connectivity: Performance analysis for the MDS-MAP algorithm,} 2010 IEEE Information Theory Workshop on Information Theory (ITW 2010, Cairo), Cairo, Egypt, pp. 1--5 (2010).
\bibitem{MNPT_PNAS} Miguel Moscoso, Alexei Novikov, George Papanicolaou, and Chrysoula Tsogka, 
Correlation-informed ordered dictionary learning for imaging in complex media
 Proc. Nat. Acad. Sci, vol. 121(11), e2314697121, 2024.
\bibitem{foldy} L. L. Foldy, The multiple scattering of waves,Phys. Rev. 67, 107-119 (1945).
\bibitem{lax}  M. Lax, Multiple scattering of waves, Rev. Mod. Phys. 23, 287-310 (1951).
\bibitem{let17}P.-D. Letourneau, Y. Wu, G. Papanicolaou, J. Garnier, and E. Darve,
{A numerical study of super-resolution through fast 3D wideband algorithm for scattering in highly-heterogeneous media,}
Wave Motion, 70, 113-134 (2017)
\bibitem{martin} Martin PA. Multiple Scattering: Interaction of Time-Harmonic Waves with N Obstacles. Cambridge University Press; 2006.
\bibitem{NW} Novikov, A., White, S. Dictionary learning for the almost-linear sparsity regime. Algorithmic learning theory, 39 pp.
Proc. Mach. Learn. Res. (2023).
\bibitem{Moscoso12} M. Moscoso, A. Novikov, G. Papanicolaou, and L. Ryzhik, {A differential equations approach to l1-minimization with applications to array imaging,} Inverse Problems {\bf 28}, 105001 (2012).

\bibitem{Moscoso14} A. Chai, M. Moscoso and G. Papanicolaou, {Imaging strong localized scatterers with sparcity promoting optimization,} SIAM Journal of Imaging Science {\bf 7}, 1358--1387 (2014) . 

\bibitem{Engan00}  K. Engan, S.O. Aase and J. Hakon Husoy, {Method of optimal directions for frame design,} in  1999 IEEE International Conference on Acoustics, Speech, and Signal Processing. 
Proceedings. ICASSP99 (Cat. No.99CH36258), {\bf 5}, 2443--2446 (1999). 
\bibitem{Shang03} Y. Shang, W. Ruml, Y. Zhang, and M. P. J. Fromherz, {Localization from mere connectivity,} in MobiHoc `03: Proceedings of the 4th ACM international symposium on Mobile ad hoc networking \& computing. New York, NY, USA: ACM, pp. 201--212 (2003).

\bibitem{Ester1996} M. Ester, H.-P. Kriegel, J. Sander, J\"{o}rg and X. Xu, {A density-based algorithm for discovering clusters in large spatial databases with noise}, in
Proceedings of the Second International Conference on Knowledge Discovery and Data Mining, AAAI Press, pp. 226-231 (1996).


\bibitem{Ioffe2015} S. Ioffe and C. Szegedy, {Batch Normalization: Accelerating Deep Network Training by Reducing Internal Covariate Shift}, in arxiv 1502.03167 (2018).



\bibitem{hutter2018} I. Loshchilov and F. Hutter, {Fixing Weight Decay Regularization in Adam}, arxiv 1711.05101 (2018)





\end{thebibliography}

%
%

\end{document}